\documentclass[letterpaper, 10 pt, conference]{ieeeconf}  

\IEEEoverridecommandlockouts                              

\usepackage{graphics} 
\usepackage{epsfig} 
\usepackage{times} 
\usepackage{amsmath} 
\usepackage{xcolor}

\usepackage{mathrsfs}
\usepackage{amssymb}
\usepackage{cite}
\usepackage{dsfont}
\usepackage{xcolor}
\usepackage{mathtools}
\usepackage{mathrsfs}
\usepackage{subfig}
\usepackage{balance}
\usepackage[ruled]{algorithm2e}
\usepackage{tcolorbox}
\usepackage{bbm}
\newtheorem{remark}{Remark}
\usepackage{fontawesome5}
\newcommand\safegeom{{{Safe-Geom}}}
\newcommand\safelang{{{Safe-Lang}}}

\usepackage{graphicx}
\usepackage{etoolbox}

\usepackage{caption}
\makeatletter
\apptocmd{\@maketitle}{\centering
\includegraphics[width=1.0\textwidth]{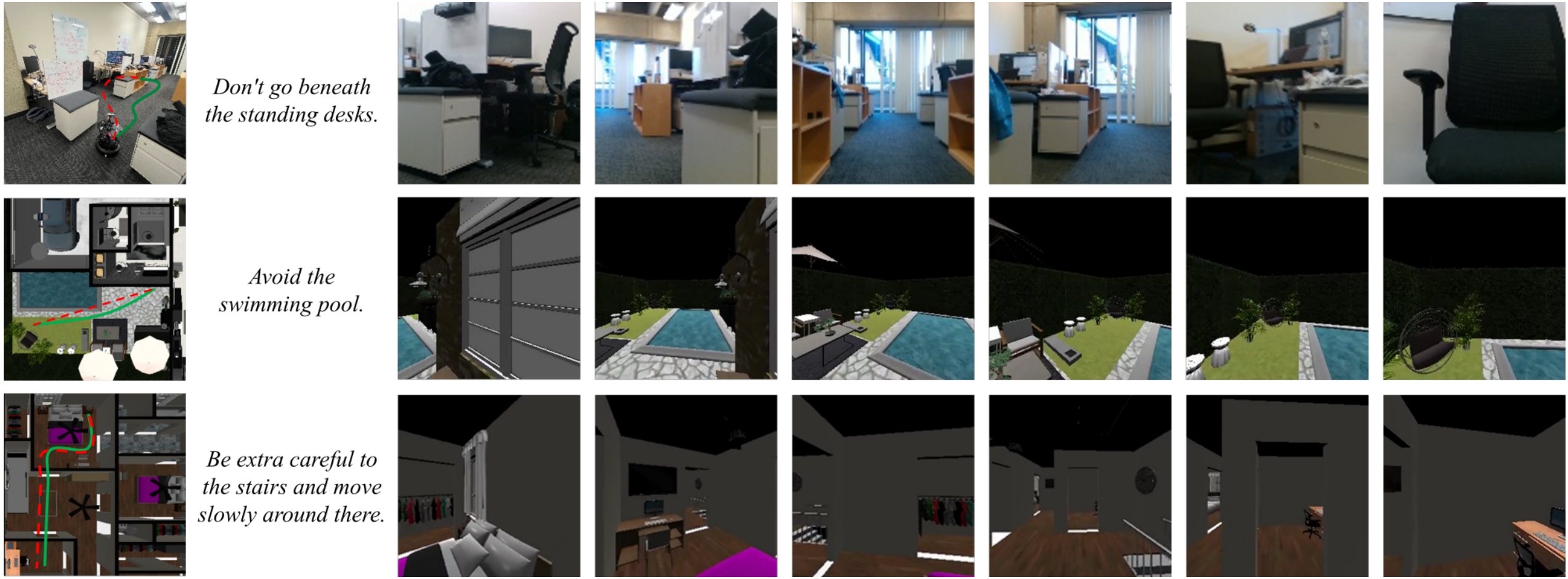}
\vspace{-0.5cm}
\captionof{figure}{Our proposed language-conditioned filter safeguards the robot against semantic constraint violations while completing navigation tasks. Red dotted lines indicate nominal (shortest) paths, and green lines denote filtered trajectories.
}
\label{fig:overview}
\setcounter{figure}{1}}{}{}
\makeatother

\title{\LARGE \bf
From Words to Safety: Language-Conditioned Safety Filtering \\for Robot Navigation
\vspace{-0.5em}
}
\author{Zeyuan Feng$^{1}$, Haimingyue Zhang$^{2}$, Somil Bansal$^{1}$
\thanks{This research is supported in part by the DARPA Assured Neuro Symbolic Learning and Reasoning (ANSR) program and by the NSF CAREER program (2240163). $^{1}$Authors are with the department of Aeronautics and Astronautics, Stanford University, Stanford, USA. $^{2}$Author is with the School of Vehicle and Mobility at Tsinghua University, Beijing, China, and  Corresponding author: \{zeyuanf@stanford.edu\}}
}

\renewcommand{\baselinestretch}{0.99}

\begin{document}

\maketitle
\thispagestyle{empty}
\pagestyle{empty}

\begin{abstract}
As robots become increasingly integrated into open-world, human-centered environments, their ability to interpret natural language instructions and adhere to safety constraints is critical for effective and trustworthy interaction. 
Existing approaches often focus on mapping language to reward functions instead of safety specifications or address only narrow constraint classes (e.g., obstacle avoidance), limiting their robustness and applicability. 
We propose a modular framework for language-conditioned safety 
in robot navigation.
Our framework is composed of three core components: (1) a large language model (LLM)-based module that translates free-form instructions into structured safety specifications, (2) a perception module that grounds these specifications by maintaining object-level 3D representations of the environment, and (3) a model predictive control (MPC)-based safety filter that enforces both semantic and geometric constraints in real time. 
We evaluate the effectiveness of the proposed framework through both simulation studies and hardware experiments, demonstrating that it robustly interprets and enforces diverse language-specified constraints across a wide range of environments and scenarios.
\end{abstract}

\section{Introduction}
\label{sec:intro}
Robots are increasingly deployed in human-centered environments such as homes, hospitals, and workplaces, where safety is fundamental for trust and adoption. 
Existing approaches predominantly address static, predefined safety constraints, most notably collision avoidance. 
Through methods such as Model Predictive Control (MPC) \cite{tearle2021predictive, bena2025geometry}, Control Barrier Functions (CBF) \cite{ames2019control, wabersich2023data}, and reachability analysis \cite{borquez2024safety, hsu2023safety}, classical techniques have proven effective at filtering unsafe actions. However, these approaches presuppose fixed safety specifications, limiting their ability to adapt to changing contexts.

In contrast, failures in human–robot interaction are often \textit{dynamic} and \textit{semantic}: they arise from context- and preference-dependent constraints that evolve with the situation and the user. We refer to this broader requirement as \textbf{semantic safety} -- the ability of robots to not only avoid collisions, but also anticipate and adapt to context-specific safety expectations expressed by humans around them. 
For example, a nurse may instruct a medical delivery robot to ``\textit{be quiet around the patient's bed}'' when there is a patient who should not be disturbed, or a homeowner may want a robot vacuum to ``\textit{not go under a standing desk}'' to avoid damage. Such constraints are difficult to anticipate and encode \textit{a priori}, yet are naturally expressed in language by non-technical users. Enabling robots to accept, ground, and enforce such natural language constraints at runtime is therefore a key missing capability.

Recent progress in vision–language models (VLMs) has primarily targeted high-level decision making, e.g., specifying tasks and goals in natural language or shaping reward functions. In contrast, work on vision-language-conditioned safety remains underexplored. Prior efforts have been limited either to policy steering or to handling a narrow class of language-specified constraints. 
Consequently, the central challenges of language-based safety remain open, i.e., (1) grounding vague or ambiguous instructions in perception, (2) generalizing across diverse instructions and constraint types, (3) maintaining interpretable, auditable specifications, and (4) enforcing them in real time during execution.

To address these challenges, we introduce a modular framework for language-conditioned safety (Fig. \ref{fig:framework_overview}). Our system integrates three components: (1) a language module that parses free-form instructions into safety configuration files using predefined constraint templates, leading to structured safety constraint specifications; (2) a perception module that persistently grounds referenced objects and regions in 3D by combining open-vocabulary and panoptic segmentation; and (3) a sampling-based safety filter that can enforce novel semantic and geometric constraints online while minimally deviating from nominal control. Together, these components enable robots to incorporate runtime language instructions into their low-level safety guarantees.
We validate our framework extensively in both simulation and hardware experiments, demonstrating robustness across diverse environments and instruction types. Ultimately, our framework leverages VLMs as ``open-world'' runtime monitors, where natural language provides a flexible interface for specifying evolving safety requirements.

\section{Related Work}
\label{Related Work}
\subsection{Large Language Models as Generalist Controllers}
Recent work has explored using LLMs as generalist reasoning engines that translate natural language into low-level plans or actions. Hierarchical systems such as SayCan \cite{ahn2022can} and PaLM-E \cite{driess2023palm} couple LLM reasoning with perception and planning modules, while RT-2 \cite{zitkovich2023rt} and OpenVLA \cite{kim2024openvla} extend this paradigm to vision-language-action policies trained end-to-end. These systems highlight the zero-shot generalization potential of LLMs, but their focus is primarily on task execution rather than enforcing explicit safety guarantees.

\begin{figure*}[t]
    \centering
    \includegraphics[width=0.9\textwidth]{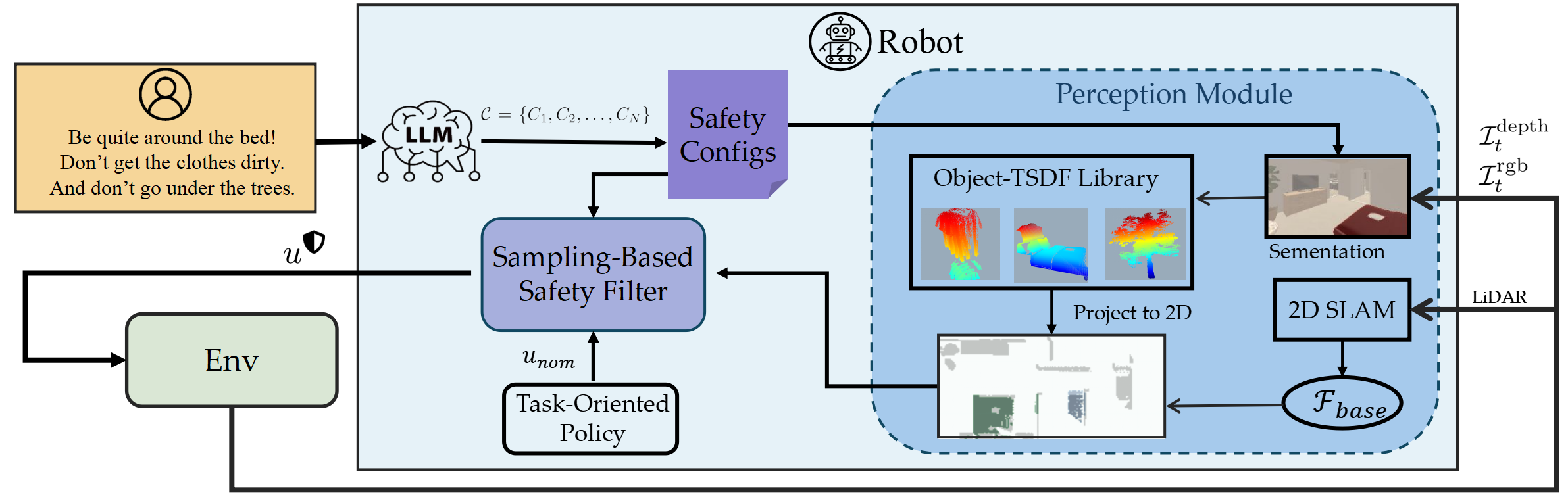}
    \caption{Framework Overview. Language instructions are parsed by an LLM into safety configurations, grounded via the perception module, and enforced by a sampling-based safety filter.}
    \label{fig:framework_overview}
    \vspace{-1.5em}
\end{figure*}

\subsection{Language for Constraints vs. Rewards}
Most language-conditioned robotics methods exploit LLM reasoning for reward design and shaping \cite{yu2023language,xie2023text2reward,goyal2019using,zeng2024learning,song2023self}, especially in reinforcement learning and navigation. 
While effective for guiding behaviors, these approaches typically treat safety as a soft cost in the optimization problem (as opposed to a strict requirement), or omit it altogether.

In contrast, only a handful of works have attempted to derive explicit safety constraints from language. For instance, \cite{wu2025foresight} aligns language with learned world models for safe policy steering, but requires heavy manual engineering and remains domain-specific. \cite{santos2025updating} integrates VLMs with Hamilton–Jacobi (HJ) reachability for semantic safety filtering, though it addresses only obstacle-avoidance-style constraints. 
\cite{brunke2025semantically} uses VLMs to infer relevant semantic constraints for a robot manipulator in a given scene, and enforces them with CBF-based filters.
Our work complements these efforts by targeting semantic safety for navigation in dynamic, unstructured environments, scaling to diverse constraint types, and supporting runtime template-based specification parsing coupled with safety filtering.

\subsection{Grounding Language in Perception}
Mapping instructions such as “\textit{be quiet around the bed}” to actionable robot constraints requires robust grounding in perception. 
Two primary approaches have been explored for providing 3D representation of the target objects: (i) semantic SLAM algorithms that construct a full 3D semantic map \cite{liao2022so, li2024sgs,zhu2024semgauss} or scene graph \cite{gu2024conceptgraphs} of the environment, and (ii) open-vocabulary segmentation models \cite{takmaz2023openmask3d, he2023clip, zhou2023zegclip} combined with camera projections, robot odometry, and point cloud fusion techniques \cite{vizzo2022vdbfusion}. 
While semantic SLAM provides detailed global representations, it is computationally expensive and prone to odometry drift in cluttered environments.
For navigation, open-vocabulary segmentation with point cloud integration is often more practical.
Our perception module adopts this latter approach, fusing CLIP-based and panoptic segmentation to balance semantic coverage with boundary accuracy.

\subsection{Safety Filtering}
Safety filtering techniques modify a robot’s nominal control in real time to ensure satisfaction of safety constraints. Approaches based on HJ reachability and CBFs \cite{wabersich2023data, ames2016control, borquez2024safety} provide formal guarantees for runtime shielding, but require precomputed safety value functions or fixed constraint sets, making them difficult to extend to dynamic, language-specified constraints.
MPC-based safety filters, by contrast, offer greater flexibility through online optimization, but their computational demands limit scalability in cluttered or dynamic settings.
To balance flexibility and efficiency, we design a predictive safety filter inspired by Model Predictive Path Integral (MPPI) control \cite{7487277,williams2018information}, which leverages parallelizable, sampling-based trajectory optimization to reduce computation while retaining the expressiveness of MPC.

\section{Problem Setup}
\label{Problem Setup}
We consider a mobile robot navigating in a static environment $\mathcal{E}$ while executing a nominal goal-directed navigation policy. The robot evolves under discrete-time dynamics $x_{k+1}=f(x_k,u_k)$, where $x \in \mathbb{R}^n$ denotes the state and $u \in \mathbb{R}^m$ is the control input.

At each time step $t$, the robot has access to: (i) a nominal control command $u_{\mathrm{nom}}$ from a task planner, (ii) RGB-D observations of the environment $(\mathcal{I}_t^{\mathrm{rgb}}, \mathcal{I}_t^{\mathrm{depth}})$, and (iii) an estimated state $\hat{x}_t$ from onboard odometry. 
The nominal control policy, $u_{\mathrm{nom}}$, is task-oriented and may be arbitrary. Importantly, this policy is not guaranteed to satisfy geometric collision-avoidance constraints or higher-level semantic safety requirements specified in natural language.

In addition, the robot may receive natural language instructions online at any time.
Specifically, the robot maintains a set of active instructions $\mathcal{L} = \{L_1, L_2, \dots, L_m\}$, which may be empty initially and evolve over time as new commands are issued.
These instructions encode semantic safety constraints $\mathcal{C}_{\mathrm{sem}}$ (e.g., ``\textit{move slowly near the crib},'' ``\textit{do not go under tables}''), which augment the basic environment-collision constraints $\mathcal{C}_{\mathrm{base}}$.
Our objective is to design a closed-loop, language-aware safety filter  
\begin{equation}
    \pi_{\mathrm{safe}} : (x_t, u_{\mathrm{nom}}, \mathcal{L}) \mapsto u_{\mathrm{filtered}},
\end{equation}
that minimally modifies $u_{\mathrm{nom}}$ to ensure compliance with both $\mathcal{C}_{\mathrm{base}}$ and $\mathcal{C}_{\mathrm{sem}}$ in real time.
The filter must satisfy three key requirements: (a) interactive operation — run at control-loop frequencies to guarantee responsiveness; (b) robustness — tolerate uncertainty and noise in perception and state estimation; and (c) generality — handle diverse categories of language-specified constraints.

\section{A Modular Framework for Handling Language Safety Instructions}
\label{sec:approach}
\noindent \textbf{Framework Overview.} We propose a modular framework for handling language-conditioned safety in mobile robot navigation (see Fig.~\ref{fig:framework_overview}).
The framework decomposes the challenging problem of semantic runtime monitoring into three tractable submodules:
\begin{itemize}
    \item a \textit{Language Module} that parses free-form safety instructions into structured constraint specifications;
    \item a \textit{Perception Module} that grounds these specifications into a semantic representation of the robot’s 3D environment; and,
    \item a \textit{Predictive Safety Filter} that enforces both semantic and geometric constraints in real time while minimally deviating from nominal control commands.
\end{itemize}
The language module is triggered asynchronously when new instructions arrive, while the perception module operates at a fixed sensing frequency. The safety filter runs synchronously at every control step, ensuring that the robot’s executed control remains safe with respect to both semantic and geometric constraints.semantic and geometric constraints.

\subsection{Safety Specification Synthesis (Language Module)}
We categorize language-specified safety constraints into three broad classes that cover the diverse types of instructions encountered in navigation tasks:
\begin{itemize}
    \item \textbf{\emph{Spatial exclusion constraints}} - prohibit the robot from entering, passing beneath, or approaching certain regions or objects (e.g., “\textit{avoid the swimming pool},” “\textit{do not go beneath tables},” “\textit{keep 0.25 m away from cars}”). These unify classic avoidance, overhead/underfoot, and buffer-margin directives.
    \item \textit{\textbf{Kinematic Modulation Constraints}} — regulate the robot’s motion either globally or within specific contexts (e.g., “\textit{reduce speed near the sofa},” “\textit{slow down on the carpet}”). These capture both global self-limitations and conditional pace-modulation rules.
    \item \textbf{\textit{Abstract Intent Constraints}} — express high-level safety guidance in vague or subjective terms (e.g., “\textit{be extra careful near stairs},” “\textit{be quiet around the bed}”). Such instructions require interpretation and are resolved into spatial exclusion or kinematic modulation primitives through language model reasoning. 
\end{itemize} 
To unify these categories, we design a JSON-style specification template:  
\[
\parbox{0.9\linewidth}{$\mathcal{T}$ = \texttt{
\{"type": str,  "obj": str, \\"buffer": float, 
"vel max": float, \\"angular vel max": float\}
}}
\]
where each field encodes the constraint type, referenced objects, buffer distance, and velocity limits.  

Each natural language instruction $L$ is translated into a structured configuration $C \in \mathcal{T}$ by an LLM prompted with the schema definition, a description of robot capabilities, and few-shot examples. The robot maintains a dynamic list of active configurations $\mathcal{C} = \{C_1, C_2, \dots, C_N\}$, capped at $M$ entries to prevent unbounded growth. When instructions are ambiguous or conflicting, the LLM is instructed to request clarification.
The resulting configurations in $\mathcal{C}$ serve as structured, interpretable inputs for both the perception module and the safety filter, as we describe later in this section.
Overall, the Language Module leverages the reasoning capabilities of LLMs to generalize beyond template-specific phrasing, enabling robust interpretation of diverse natural language instructions. 
By keeping the parsing process asynchronous from the control loop, the framework ensures that language inputs -- often sparse and low-frequency -- are transformed into structured, actionable specifications without compromising real-time safety enforcement.
It is also important to note that as VLMs continue to advance, our modular framework can naturally incorporate additional classes of constraints by defining new schema fields, thereby extending beyond the three categories considered here.

\subsection{Constraint Grounding (Perception Module)}
Once the set of safety specifications $\mathcal{C}$ is generated by the language module, the perception module grounds each specification $C_i$ into a semantic failure set $\mathcal{F}_{\mathrm{sem}}^{i}$ for downstream safety filtering.
Our design emphasizes computational and memory efficiency, avoiding the overhead of full 3D semantic SLAM while still enabling persistent, semantically grounded object tracking.

\noindent \textbf{Semantic Mask Generation:} We adopt a hybrid segmentation strategy that combines the strengths of open-vocabulary and panoptic segmentation. CLIP-based segmentation \cite{lueddecke22_cvpr} provides open-vocabulary recognition but coarse object boundaries, while panoptic segmentation yields accurate object boundaries but lacks semantic label assignment. 
For each safety specification and RGB frame, we first compute both segmentation outputs, then fuse them by only retaining panoptic regions whose intersection-over-union (IoU) with the CLIP mask exceeds a predefined threshold. 
This produces accurate, vocabulary-flexible semantic masks, denoted $M_t^i = \operatorname{Seg}(\mathcal{I}_t^{\mathrm{rgb}}, C_i)$.

\noindent \textbf{Localization and Mapping:} We assume that a lightweight 2D LiDAR-based SLAM module maintains a locally consistent occupancy grid for basic collision avoidance $\mathcal{F}_{\mathrm{base}}$ and provides accurate robot pose estimates. Linear and angular velocities are obtained from onboard odometry.  

\noindent \textbf{Point Cloud Integration:} 
For each $C_i$, RGB-D depth data within mask $M_t^i$ are projected into a 3D point cloud using camera intrinsics and robot pose:
\begin{equation}
    \operatorname{proj}\left(\mathcal{I}_t^{\mathrm{depth}}, M_t^i, x\right) \rightarrow \mathbf{p}_t^i.
\end{equation}
The resulting $\mathbf{p}_t^i$ is integrated into a voxel-based truncated signed distance function (TSDF) representation $\mathbf{p}^i$, enabling persistent object tracking associated with $C_i$. Each object-level TSDF is then projected into a 2D occupancy map $\mathcal{O}_i$ to support constraint evaluation.

\noindent \textbf{Grounding semantic failure set:} Finally, each specification $C_i$ is mapped to a semantic failure set $\mathcal{F}_{\mathrm{sem}}^{i} = \left\{l_i(x) - \text{buffer}_i\leq 0 \right\}$, where $l_i(x)$ is a constraint-specific scalar function. 
For instance:
\begin{equation}
\begin{aligned}
\text{Spatial Exclusion:} &\quad l_i(x) = \operatorname{SDF}(x; \mathcal{O}_i), \\
\text{Kinematic Modulation:} &\quad l_i(x) = \min(v_{\max}^i - v, \omega_{\max}^i - \omega), \\
\text{Intent constraints:} &\quad l_i(x) = \max\left(\operatorname{SDF}(x; \mathcal{O}_i), \right. \\
& \qquad \quad \left. \min(v_{\max}^i - v,\omega_{\max}^i - \omega)\right).
\end{aligned}
\end{equation}
The overall semantic failure set is then
\begin{equation}
    \mathcal{F}_{\mathrm{sem}} = \cup_{i=1}^{N} \mathcal{F}_{\mathrm{sem}}^i = \left\{min_i (l_i(x) - \text{buffer}_i)\leq 0 \right\}.
\end{equation}

\subsection{Safety Filtering}
The nominal control command $u_t^{\mathrm{nom}}$ is generated by a task planner that is agnostic to safety constraints. To ensure constraint satisfaction at runtime, we introduce two sampling-based, MPC-inspired safety filters -- analogous to least-restrictive and smooth-blending safety filters \cite{borquez2024safety} -- that enforce safety while minimally deviating from the nominal control. 
Compared to existing CBF-based or HJ reachability-based filters, which rely on precomputed value functions or fixed constraint sets, the proposed sampling-based predictive filters offer greater flexibility for handling dynamically generated language instructions and heterogeneous constraint types (e.g., geometric obstacles, velocity bounds, or hybrid combinations). 
Both filters share a common backbone: a safety score computation and a control filtering policy.
Intuitively, at every step, we estimate how “safe” the future looks by simulating possible trajectories (safety score), and then select or adjust the control input (filtering policy) to maximize safety while preserving task performance.

\noindent \textbf{Safety Score Computation.}
Given an initial state $x_{\mathrm{init}}$, the safety score is computed by solving the following finite-horizon optimization problem:
\begin{equation}
\label{eq:safety_prob}
    \begin{aligned}
V^{\mathrm{safe}} & =\max_{u_{0: H-1}}  \min_{k\in \left\{ 0, 1, \ldots ,H\right\}} l\left(x_{k}\right)\\
\text { s.t. } & x_{k+1}=f\left(x_{k}, u_{k}\right),~ u_{k} \in \mathcal{U},~ x_{0}=x_{\mathrm{init}},
\end{aligned}
\end{equation}
where $l(x) := \min(l_{\mathrm{sem}}(x), l_{\mathrm{base}}(x))$ encodes the combined semantic and geometric safety constraints. 
Its sub-zero level set corresponds to the overall failure set, i.e., $\mathcal{F}_{\mathrm{base}} \cup \mathcal{F}_{\mathrm{sem}} = \{x : l(x) \leq 0\}$. 
Intuitively, $V^{\mathrm{safe}}$ represents the maximum worst-case safety margin achievable over the horizon.
Thus, the optimization seeks the trajectory with the highest safety margin over the horizon. 
The optimal control at $x_{\mathrm{init}}$ is given by $u^{\text{\faShield*}} := u_0^*$.
If $V^{\mathrm{safe}} \leq 0$, no safe backup trajectory exists.

In this work, we approximate this optimization via a sampling-based MPC procedure (Alg.~\ref{alg:sampling_mpc}). At each iteration, $N$ control sequences are sampled around the current best candidate, rolled out over horizon $H$, and scored by their minimum safety margin. The “safest” sequence is then used to warm-start the next iteration. This iterative refinement process guarantees a non-decreasing safety score across $R$ iterations and returns the best trajectory discovered.
\begin{algorithm}[h!]
\caption{Sampling MPC-Based Safety Score Estimation (SB-MPC)}\label{alg:sampling_mpc}
\textbf{Input:} initial state $x_{\mathrm{init}}$, initial guess of the control sequence $\mathbf{u}$\;
\textbf{Output:} $V^{\mathrm{safe}}$, safest control sequence found $\mathbf{u}^*$, the optimal safe control $u^{\text{\faShield*}}$ \;
\textbf{Parameters:} number of iterative refinement steps $R$, number of control samples $N$, sampling variance $\sigma$;
    \ForEach{$r=1:R$}{
        \ForEach{$n=1:N$}{
            $\tilde{\mathbf{u}}^n \sim \text{Gaussian}(\mathbf{u}, \sigma^2)$ \;
            $x^n_0 \leftarrow x_{\mathrm{init}}$ \;
            \For{$h=0:H-1$}{
        $x^n_{h+1}= f(x^n_{h},\tilde{\mathbf{u}}^n_h)$ \;
            }
            $J^n= \min_{h=0,1,...,H} l(x^n_{h})$ \;
        }
        $n^*=\arg\max_n J^n $ \;
        $\hat{V} \leftarrow  J^{n^*} $ \;
        $\mathbf{u} \leftarrow \tilde{\mathbf{u}}^{n^*}$ \;
    }
    \textbf{Return} $\mathbf{u}^* \leftarrow \mathbf{u}$, $u^{\text{\faShield*}} \leftarrow \mathbf{u}_0$, $V^{\mathrm{safe}} \leftarrow \hat{V}$ \;
\end{algorithm}
\remark{For navigation tasks, the effective value function typically converges within a short horizon because the robot can brake to a stop. Hence, a relatively small $H$ suffices for real-time computation.}
\begin{figure*}[h!]
    \centering
    \includegraphics[width=0.95\textwidth]{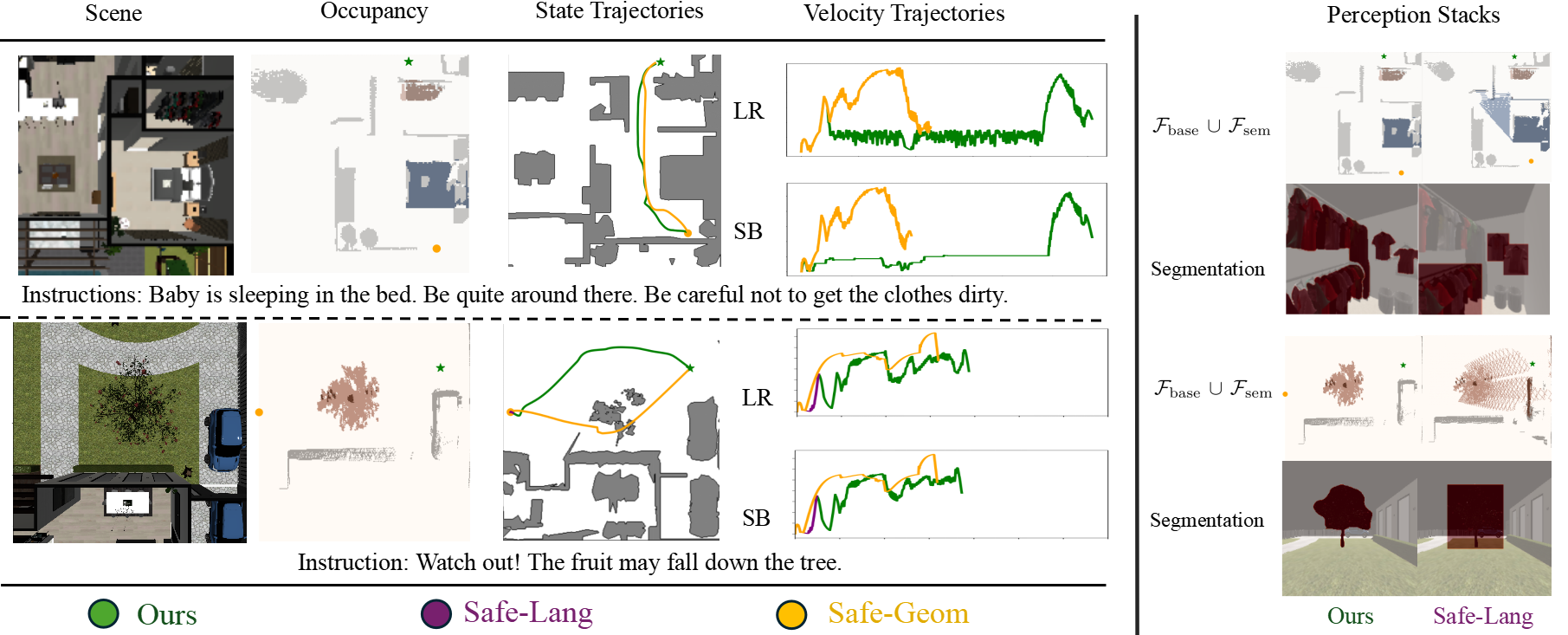}
    \caption{Qualitative results. Left: our method follows language instructions and reaches the goal, while Safe-Geom violates semantic constraints. Right: Safe-Lang tends to produce overly conservative semantic failure sets, hindering task completion.}
    \label{fig:qualitative_results}
    \vspace{-1.5em}
\end{figure*}

\noindent \textbf{Least-Restrictive Safety Filter.}
The least-restrictive filter executes the nominal control whenever its predicted trajectory maintains a positive safety margin above a threshold $\epsilon$. Otherwise, it switches to the backup safe control obtained from SB-MPC in Alg.~\ref{alg:sampling_mpc}.
Specifically, let $V^{\mathrm{safe}}_{\mathrm{nom}}$ denote the safety score of the nominal controller $u_{\mathrm{nom}}^k$ at the current state $x_k$. The overall filtered control is given by:
\begin{equation}
    u_{\mathrm{filtered}} = \begin{cases}
        u_{\mathrm{nom}}^k,~~\mathrm{if}~~ V^{\mathrm{safe}}_{\mathrm{nom}} > \epsilon\\
        u^{\text{\faShield*}},~~\mathrm{otherwise}  
    \end{cases}
\end{equation}
where $u^{\text{\faShield*}}$ is the optimal safe control obtained from SB-MPC in Alg.~\ref{alg:sampling_mpc}.  
\remark{Maintaining a warm-start sequence $\mathbf{u}^{\mathrm{init}}$ is essential for avoiding sampling failures in marginally safe states. Without a good initialization, stochastic sampling may miss safe trajectories even when they exist. We use the last optimal control sequence obtained from SB-MPC as the initial guess.}

\remark{Because of the finite horizon and stochasticity, recursive safety is not formally guaranteed under the proposed filtering method. Empirically, however, navigation safety remains high: the robot can always stop within a few steps, and distant obstacles beyond the horizon have a negligible effect.}

\noindent \textbf{Smooth-Blending Safety Filter.}
The smooth-blending filter aims for minimal deviation from nominal behavior while maintaining safety. Rather than switching abruptly to a backup trajectory, it searches locally around $u^k_{\mathrm{nom}}$ for the closest safe control, blending task performance with safety guarantees. If no nearby safe control exists, it falls back to the fully safe SB-MPC trajectory. 

Specifically, we begin with a library of primitive control actions $\mathcal{U}_{\mathrm{prim}}$. For each candidate action $u_{\mathrm{prim}} \in \mathcal{U}_{\mathrm{prim}}$, we evaluate its safety score $V^{\mathrm{safe}}_{\mathrm{prim}}$.  We then retain only those actions that satisfy the CBF-inspired blending condition: $V^{\mathrm{safe}}_{\mathrm{prim}} - V^{\mathrm{safe}} \geq -\gamma V^{\mathrm{safe}}$, which ensures that the decrease in safety margin under the new action is bounded relative to the current safety score.
Intuitively, this means that when the system is already near the safety boundary ($V^{\mathrm{safe}} \approx 0$), the filter prevents actions that would reduce the margin further. Conversely, when the system is well within the safe set (large $V^{\mathrm{safe}}$), the filter allows more flexibility to prioritize task performance.
Finally, among the filtered primitive actions, we pick the control action that is closest to the nominal control input.
This blending strategy produces smoother task execution, avoids abrupt switching to backup trajectories, and has shown improved robustness to model mismatch in real-world experiments \cite{borquez2024safety}.

\remark{In both the least-restrictive and the smooth-blending filter, safety score computations for all candidate states can be parallelized, which dramatically speeds up the filtering process.}

\section{Simulation Results}
\label{sec:exp_setup}
\begin{figure*}[h!]
    \centering
    \includegraphics[width=\textwidth]{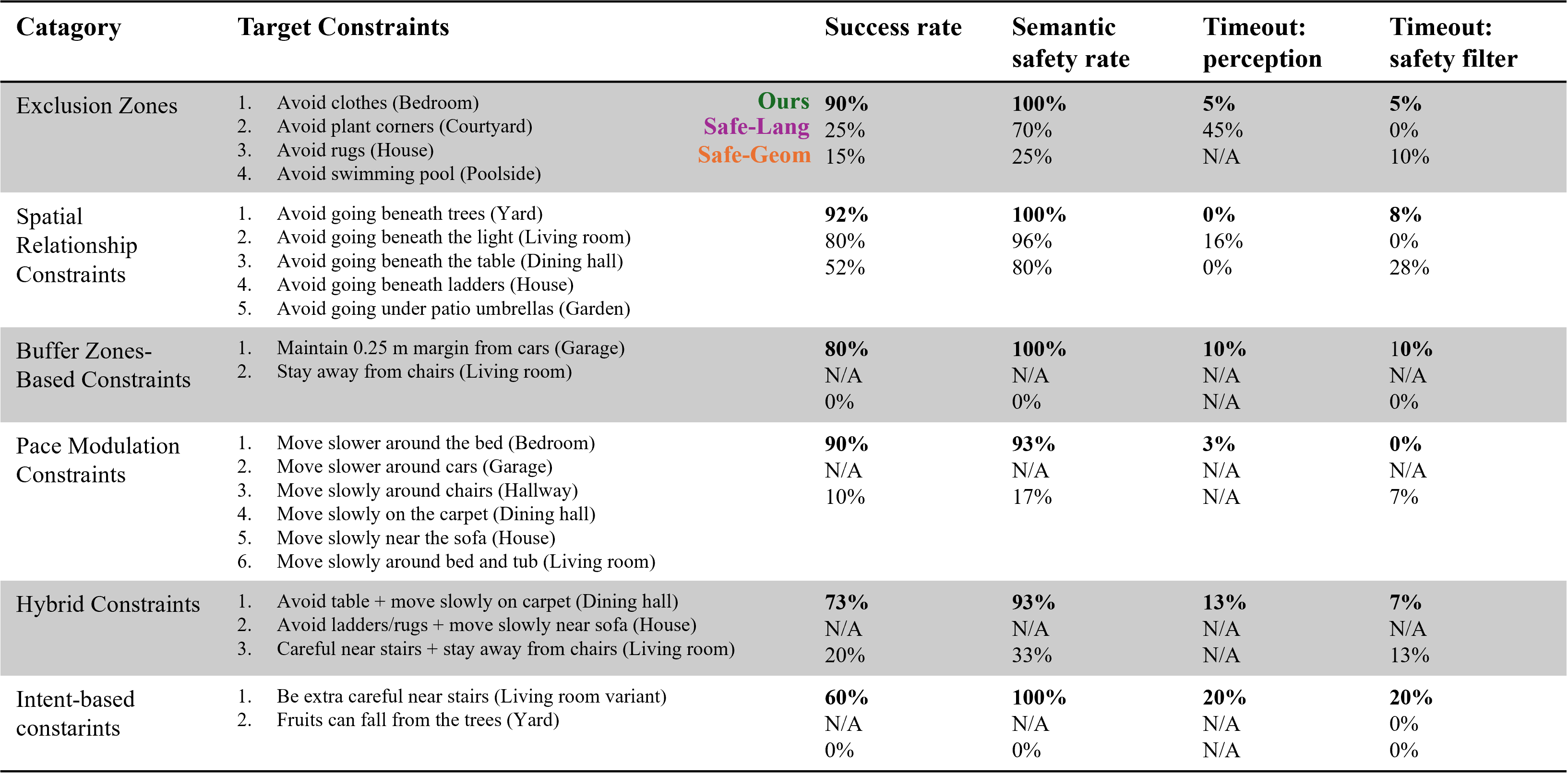}
    \caption{Quantitative simulation results. Our method demonstrates broad generality and consistently achieves the best performance across all baselines.}
    \label{fig:quantitative_results}
    \vspace{-1.5em}
\end{figure*}
\noindent \textbf{Robot Dynamics.} We evaluate our framework in both simulated and hardware experiments. The robot is modeled as a 4D Dubins car with $
x = [p_x, p_y, \theta, v]^\top \in \mathcal{X} \subseteq \mathbb{R}^4$,
where $(p_x, p_y)$ denotes planar position, $\theta$ the heading angle, and $v$ the linear velocity. The control input is $u_t = [\omega, a]^\top \in \mathcal{U} \subseteq \mathbb{R}^2$, where $\omega$ and $a$ represent bounded angular velocity and linear acceleration, respectively. The discrete-time dynamics with a time step of $\Delta$ are given by:
\begin{equation} \label{eqn:dynamics_sim}
\begin{aligned}
p_{x,t+1} &= p_{x,t} + v_t \cos\theta_t \Delta, \\
p_{y,t+1} &= p_{y,t} + v_t \sin\theta_t \Delta, \\
\theta_{t+1} &= \theta_t + \omega_t \Delta, \\
v_{t+1} &= v_t + a_t \Delta.
\end{aligned}
\end{equation}
We impose bounds to ensure physically realistic motion: $0 \leq v_t \leq v_{\max}$, where $v_{\max}$ is set to $1.5 \,\mathrm{m/s}$.
Angular velocity and acceleration are bounded by $|\omega| \leq 1 \,\mathrm{rad/s}$ and $|a| \leq 3 \,\mathrm{m/s^2}$, respectively.

\noindent \textbf{Simulation Environment.}
We use the Habitat 3.0 simulator to evaluate our framework in diverse indoor and outdoor environments. We selected 12 environments with associated language instructions, covering:  
1) \emph{Home-like scenes} with furniture, rugs, and narrow corridors;  
2) \emph{Public or semi-open spaces} such as dining halls with wide hallways and sparse obstacles;  
3) \emph{Specialized scenarios} such as yards with swimming pools or gardens with trees, chosen to test nuanced safety instructions.  
For each environment, we test the robot on several language instructions that evaluate the ability of the proposed framework to respect exclusion zones, spatial relationship constraints, pace modulation constraints,  intent-based constraints, and hybrid constraints (see Fig. \ref{fig:quantitative_results} for a list of constraints). 
For each language instruction, we evaluate on five randomized trials.
The language instructions are all issued at $t = 0$, and each trial is terminated after 35 seconds. 
Finally, in simulation, we assume perfect state estimation and noise-free RGB-D and LiDAR inputs.

\noindent \textbf{Implementation Details.}
In all experiments, we use GPT-4o with a fixed initial prompt to parse language instructions into structured safety constraints. For perception, we use mask2former \cite{cheng2021mask2former} for panoptic segmentation and ClipSeg \cite{lueddecke22_cvpr} for open-vocabulary object recognition.
We use $N=50$ and $R=2$ for SB-MPC-based safety score estimation. For least-restrictive filtering, we use $\epsilon = 0.25$, which corresponds to the robot's base radius. For the smooth blending filter, we use $\gamma = 5.0$ and 25 control primitives.

\noindent \textbf{Nominal Policy.}
Our nominal policy consists of an A* planner, which generates the shortest path $\tau$ based on $\mathcal{F}_{\mathrm{base}} \cup \mathcal{F}_{\mathrm{sem}}$, followed by a low-level PD tracking controller without any safety penalty. The proposed safety filter wraps around the nominal control to enforce safety at runtime. For all simulation and hardware experiments, the A* planner is operating at 2~Hz while the PD controller is at 20~Hz.   

\noindent \textbf{Baselines.}
We compare our approach with:  
1) \textbf{\safegeom}: A geometric baseline where nominal actions are filtered to ensure safety with respect to $\mathcal{F}_{\mathrm{base}}$.  
2) \textbf{\safelang}~\cite{santos2025updating}: A method that updates safety representations using OWLv2 for open-vocabulary segmentation and maintains safety against both the semantic and geometric failure sets.
To enable comparison, we replace its original HJ-reachability-based safety filter with our MPC-based safety filter.

\noindent \textbf{Evaluation Metrics.}
We evaluate performance using four metrics. \emph{Success rate} measures the percentage of scenarios in which the robot reaches the goal while satisfying both geometric and semantic safety constraints. \emph{Semantic safety rate} captures the fraction of scenarios where semantic constraints are respected throughout the entire trajectory. We also report two types of failures: \emph{Timeout:perception}, which occurs when overly conservative perception enlarges the semantic failure set and prevents goal completion, and \emph{Timeout:safety filter}, which arises when the safety filter persistently overrides the nominal policy, causing failure to reach the goal despite available feasible paths.

\subsection{Overall Results} 
Results are summarized in Fig.~\ref{fig:quantitative_results}, classified by constraint types. Each constraint was evaluated over five runs, and average metrics are reported. \textbf{Safe-Geom} cannot explicitly reason about semantics, resulting in a very low semantic safety rate and consequently a low overall success rate. \textbf{Safe-Lang} can handle exclusion-zone constraints but frequently generates overly conservative masks, which block feasible paths and lead to high timeout rates and poor task completion. This effect is evident in Fig.~\ref{fig:qualitative_results}, where Safe-Lang over-segments the tree and eliminates safe passages. In contrast, our approach produces more accurate segmentation and robust occupancy integration, maintaining safety while avoiding unnecessary restrictions on motion.

In terms of other types of semantic constraints, while \safelang~lacks the capability to directly enforce spatial relationship constraints, such as “beneath”, it projects the bounding box for the object specified in the instruction down to the ground, which inadvertently allows it to avoid such relational constraints. However, \safelang, once again, leads to an overly conservative bounding box in this case, leading to a higher timeout rate compared to the proposed approach.
Moreover, it is not able to handle other types of semantic constraints, e.g., based on buffer zones, pace modulation, and intent-based constraints.
In contrast, our method is consistently able to maintain a high semantic safety rate, showing the effectiveness and generality of the proposed approach. 
Overall, these results highlight that our framework balances semantic safety compliance with efficient task execution more effectively than existing approaches.

We further observe that the \textbf{smooth-blending} filter is particularly effective for handling pace-modulation constraints (e.g., “move slowly near the bed”), yielding smooth velocity profiles, while showing smaller differences compared to the least-restrictive variant in exclusion-zone tasks (e.g., “avoid the tree”). We also note that this advantage comes at a cost of higher latency compared to the least-restrictive filter (10 ms vs 2 ms on average), though both filters remain real-time.
\begin{figure}[h!]
    \centering
    \includegraphics[width=0.9\linewidth]{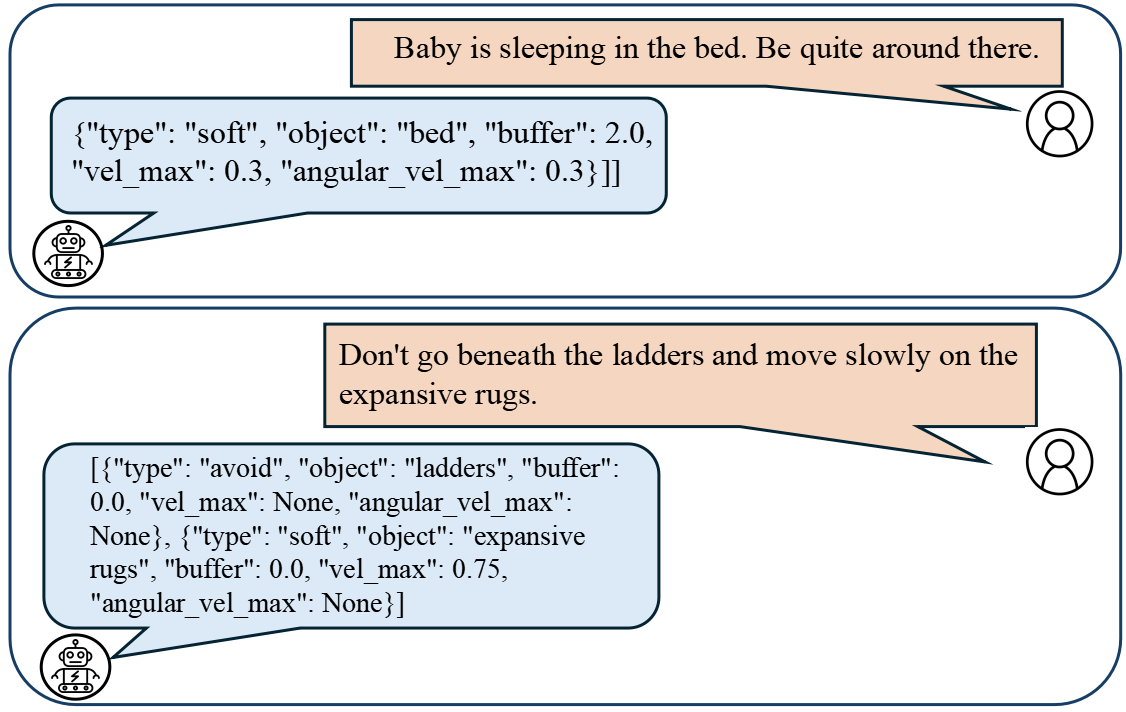}
    \caption{Representative outputs of safety specifications from the language module.}
    \label{fig:llm_quality}
    \vspace{-1.5em}
\end{figure}
\subsection{LLM Parser Evaluation}
\begin{figure}[h!]
    \centering
    \includegraphics[width=0.8\linewidth]{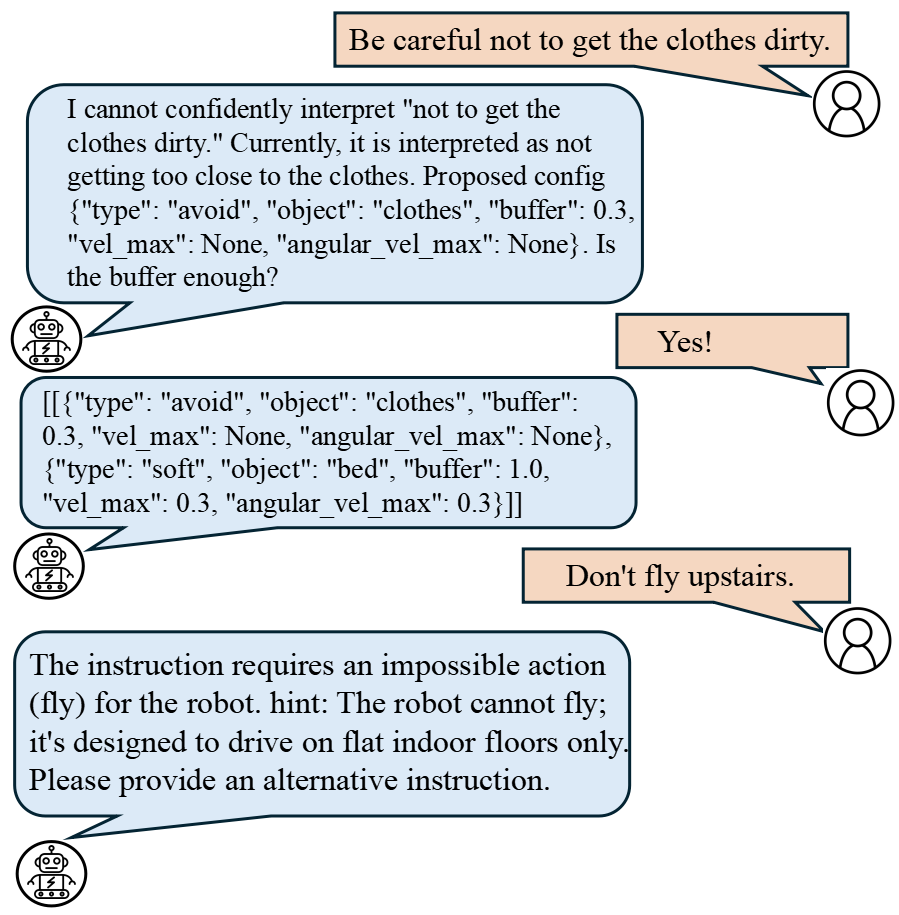}
    \caption{An illustrative example of human–LLM interaction for ambiguous instructions.}
    \label{fig:llm_interaction}
    \vspace{-1.3em}
\end{figure}

We evaluate the performance of GPT-4o in parsing natural language safety instructions into structured constraints. The LLM is given a fixed initial prompt specifying the task. For each of the 12 simulation instructions, we queried GPT-4o five times. The model successfully produced correct safety constraints in 52 out of 60 attempts without requiring additional clarification, while the remaining cases prompted the user for clarification. Fig.~\ref{fig:llm_quality} illustrates representative outputs. Notably, the LLM parser is able to translate relatively ambiguous instructions into correct formal constraints in a zero-shot manner (with only the initial prompt).  
When instructions are too ambiguous or underspecified, users can interactively refine the query to ensure that the parsed constraints match their intent in real time.
An example of this is shown in Fig.~\ref{fig:llm_interaction}.

\section{Hardware Experiments}
\label{sec:results}
\begin{figure}[h!]
    \centering
    \includegraphics[width=0.95\linewidth]{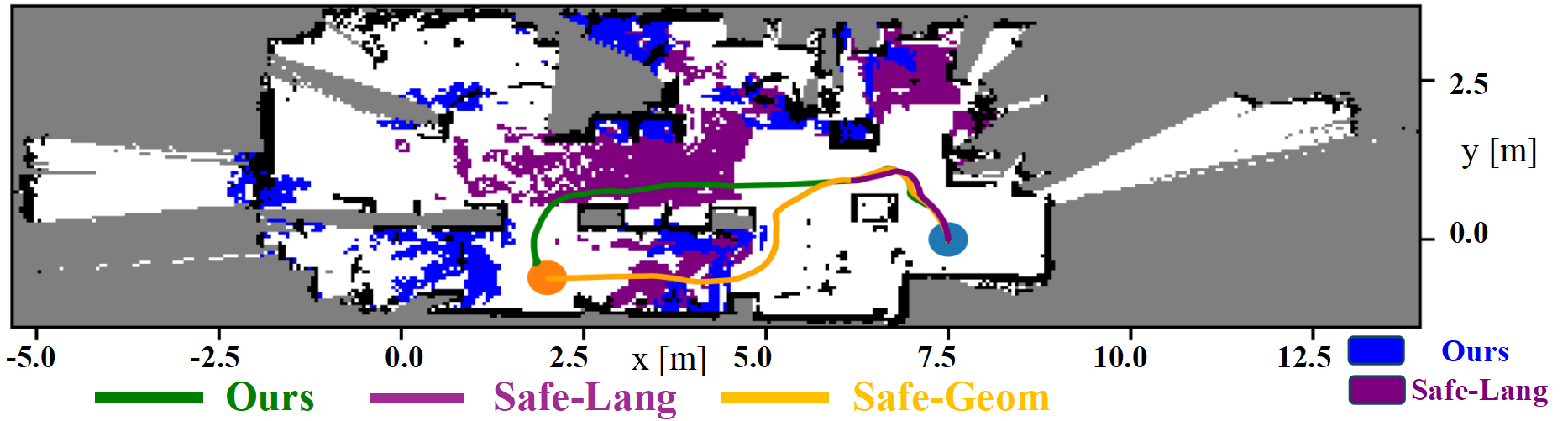}
    \caption{Hardware experiment in a cluttered office. The robot reaches the goal while respecting semantic constraints with our method. Semantic failure sets from our method and Safe-Lang are shown in blue and purple, respectively.}
    \label{fig:hardware_result}
    \vspace{-1.5em}
\end{figure}
We deployed our framework on a TurtleBot 2 platform equipped with an Intel RealSense D415 RGB-D camera and a 2D LiDAR for localization and mapping. 
The robot dynamics used for safety filtering and MPC follow \eqref{eqn:dynamics_sim}, with maximum forward velocity capped at $v_{\max} = 0.5  \mathrm{m/s}$.

We evaluate our method and \safegeom~in a cluttered office with chairs, desks, and monitors, with its representative snapshot shown in Fig.\ref{fig:overview}.
The user issues the constraint $L =$ “\textit{Don’t go under the standing desk}” at deployment time ($t = 0$). This instruction requires reasoning about a spatial relationship (avoiding going beneath a structure), which is not explicitly modeled in the geometric SLAM map.
As a result, while the nominal planner (\safegeom) successfully avoided static obstacles and reached the goal, it violated the semantic constraint by attempting to pass under the desk to take a shorter route to its goal.
\safelang~leads to an overly conservative semantic failure set in this case, leading to a timeout.
In contrast, our method successfully avoided both physical and semantic constraints, achieving safe goal-directed navigation by filtering the nominal control through its language-conditioned safety module.
This illustrates the value of modular, language-aware safety filtering in bridging the gap between human intent and real-world robot execution.

It is also worth noting that the highly cluttered nature of this environment led to odometry drift and noisy depth sensing due to occlusions and non-Lambertian surfaces (e.g., reflective monitor screens). Despite these challenges, our approach maintained a reliable grounding of the “desk” constraint and enforced it consistently throughout execution. This further highlights the robustness of the perception module and its integration with the safety filter.

\section{Limitations and Future Work}
\label{sec:conclusion}
In this work, we introduced a modular framework for language-conditioned safety in robot navigation, integrating (i) an LLM-based specification synthesis module, (ii) a perception module for semantic grounding, and (iii) a sampling-based safety filter for real-time enforcement. Through simulation and hardware experiments, we demonstrated that this framework can robustly interpret diverse natural language instructions, ground them into actionable safety constraints, and enforce them during execution, even in cluttered and noisy environments.

Despite these promising results, several limitations remain. First, failure modes can arise from instructional ambiguity, missed object detections, or odometry drift, which propagate into incorrect safety set definitions. Another limitation arises from overly conservative segmentation of the relevant object, leading to the robot freezing problem.
 
Second, the current safety filter is limited to discrete-time guarantees and assumes deterministic dynamics, which may not fully capture the uncertainties present in real-world operation. Extending the safety filter to reason about dynamics uncertainty and probabilistic occupancy updates (e.g., from noisy depth sensors) would provide stronger guarantees under real-world disturbances.

Third, latency in perception and filtering can degrade responsiveness, particularly when new obstacles suddenly enter the environment. 
Finally, while this work focused on mobile robot navigation, manipulation tasks are inherently more semantic and safety-critical in service robotics. Extending language-conditioned safety to physical interaction—where constraints may involve fragile objects, force limits, or user preferences—presents a nontrivial but important next step.


\bibliographystyle{IEEEtran}
\bibliography{arXiV/references}

\end{document}